\documentclass[10 pt, conference]{IEEEtran}
\IEEEoverridecommandlockouts

\usepackage{fancyhdr}

\usepackage[noadjust]{cite}

\usepackage{amsmath}
\usepackage{graphicx}
\usepackage{tabularx}
\usepackage{textcomp}
\usepackage{xcolor}
\usepackage{balance}
\usepackage{comment}
\def\BibTeX{{\rm B\kern-.05em{\sc i\kern-.025em b}\kern-.08em
    T\kern-.1667em\lower.7ex\hbox{E}\kern-.125emX}}
\usepackage[english]{babel}
\usepackage[utf8]{inputenc}
\usepackage{algorithm}
\usepackage[noend]{algpseudocode}
\usepackage{subcaption}
\usepackage[font = small]{caption}
\usepackage{multirow}
\usepackage{graphics}
\usepackage{adjustbox}
\usepackage{soul}

\usepackage{booktabs}
\usepackage{graphicx}
\begin{document}

\title{FlatENN: Train \underline{Flat} for \underline{E}nhanced Fault Tolerance of Quantized Deep \underline{N}eural \underline{N}etworks}

\makeatletter
\newcommand{\linebreakand}{%
  \end{@IEEEauthorhalign}
  \hfill\mbox{}\par
  \mbox{}\hfill\begin{@IEEEauthorhalign}
}
\makeatother

\author{\IEEEauthorblockN{Akul Malhotra}
\IEEEauthorblockA{
\textit{Purdue University}\\
West Lafayette, Indiana \\
malhot23@purdue.edu}

\and
\IEEEauthorblockN{Sumeet Kumar Gupta}
\IEEEauthorblockA{
\textit{Purdue University}\\
West Lafayette, Indiana \\
guptask@purdue.edu}

}







\maketitle

\pagestyle{plain}

\begin{abstract}
Model compression via quantization and sparsity enhancement has gained an immense interest to enable the deployment of deep neural networks (DNNs) in resource-constrained edge environments. Although these techniques have shown promising results in reducing the energy, latency and memory requirements of the DNNs, their performance in non-ideal real-world settings (such as in the presence of hardware faults) is yet to be completely understood. In this paper, we investigate the impact of bit-flip and stuck-at faults on \textit{activation-sparse} quantized DNNs (QDNNs). We show that a high level of activation sparsity comes at the cost of larger vulnerability to faults. For instance, activation-sparse QDNNs exhibit up to 11.13\% lower accuracy than the standard QDNNs. We also establish that one of the major cause of the degraded accuracy is sharper minima in the loss landscape for activation-sparse QDNNs, which makes them more sensitive to perturbations in the weight values due to faults. Based on this observation, we propose  the mitigation of the impact of faults by employing a sharpness-aware quantization (SAQ) training scheme. The activation-sparse and standard QDNNs trained with SAQ have up to 19.50\% and 15.82\% higher inference accuracy, respectively compared to their conventionally trained equivalents. Moreover, we show that SAQ-trained activation-sparse QDNNs show better accuracy in faulty settings than standard QDNNs trained conventionally. Thus the proposed technique can be instrumental in achieving sparsity-related  energy/latency benefits without compromising on fault tolerance. 
\end{abstract}


\vspace{-0.1in}

\section{Introduction}
\label{sec:intro}

The remarkable success of deep neural networks (DNNs) for tasks involving decision-making and sensory processing \cite{coca} \cite{go} has prompted the exploration of DNN accelerator designs for various applications \cite{dnn_accelerator}. However, the performance benefits of state-of-the-art DNNs come at the cost of large storage and computation requirements, introducing various design challenges, especially for energy- and memory-constrained edge applications. The need to reduce the size and computational complexity of DNNs has led to the emergence of Quantized Deep Neural networks (QDNNs). Various quantization techniques based on reduced bit precision of the weights and activations of DNNs have been proposed to achieve energy savings associated with storage, computation and communication, while alleviating the accuracy drop with respect to their full-precision counterparts \cite{LQNet}\cite{google_quant}.  

\begin{figure}[t!]
\centering
  \includegraphics[width = \linewidth]{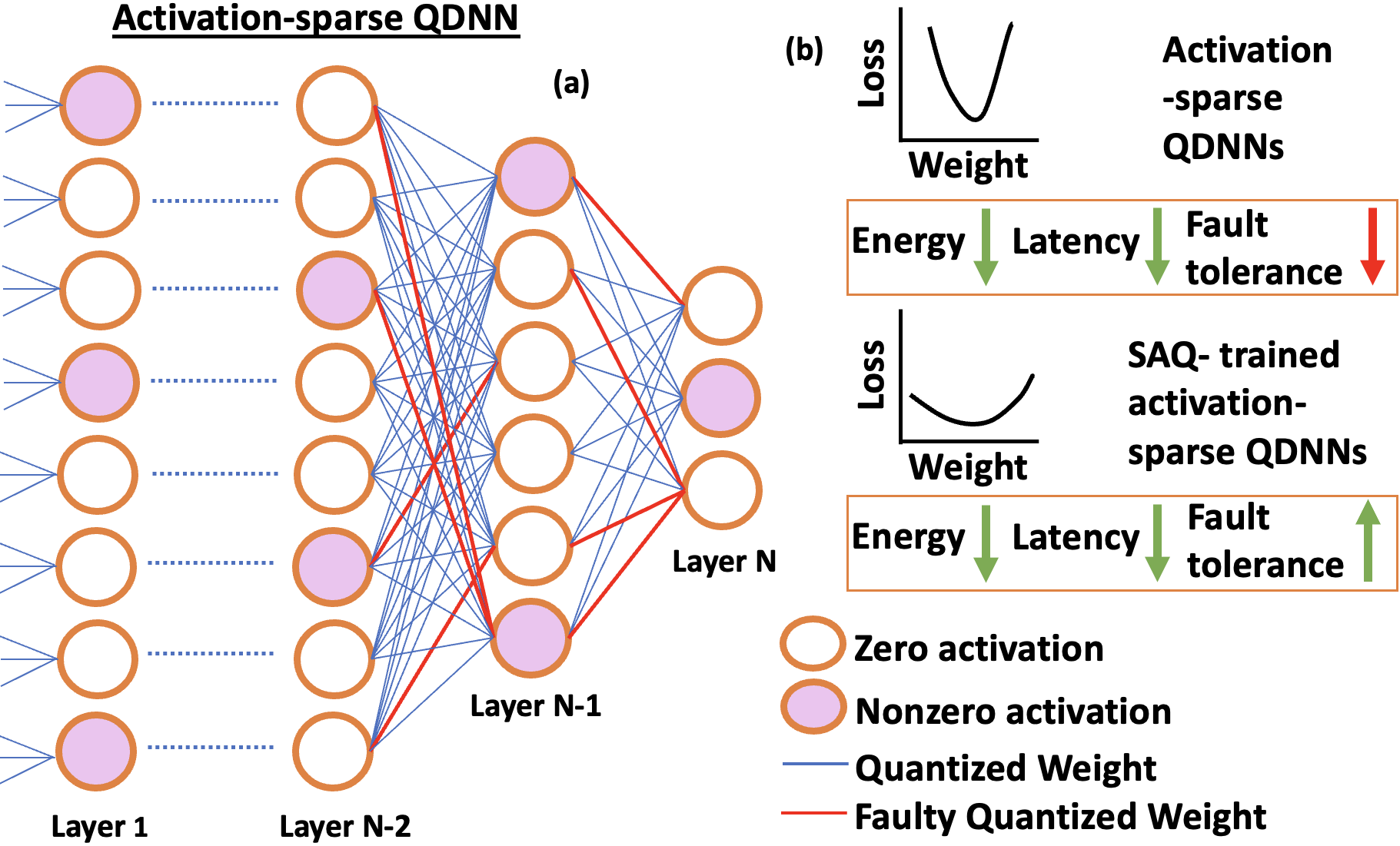}
  \caption{(a) Shows an activation-sparse (AS) QDNN in a faulty scenario. We show in our work that AS QDNNs are more prone to severe accuracy degradation than their standard counterparts. (b) describes the trade-off between the energy/latency benefits due to enhanced AS and reduced fault tolerance. We propose a fault mitigation strategy which enables AS QDNNs to retain fault tolerance by flattening its weight loss landscape using sharpness-aware quantization (SAQ) based training.}
  \label{fig:overview}
\end{figure}

Another popular method to reduce the resource requirements of DNNs is  to sparsify the weights and activations by systematically removing the less important portions of the network. Various techniques to induce weight sparsity, also referred to as weight pruning, have led to significant compression of DNNs with little or no loss of accuracy \cite{pruning1} \cite{pruning2}. Recently, approaches to increase and leverage the sparsity in the activations have also gained attention to reduce the memory requirements and improve inference speed \cite{act_sparse1} \cite{act_sparse2}. Such techniques involving weight and activation sparsification can be used in conjunction with quantization to design  QDNNs optimized for edge applications. 

While sparsified QDNNs offer promising attributes for resource-constrained systems, their deployment in real world settings also needs to consider the non-ideal hardware behavior.  DNN accelerator weights are generally stored on-chip in the memory which experiences various types of faults such as stuck-at and bit-flip faults. These faults corrupt the weight values and degrade system accuracy, Although the impact of faults in conventional DNN accelerators is well understood \cite{dnn_faults1} \cite{dnn_faults2}, the fault tolerance of weight/activation sparse DNNs has been studied only to a limited extent. Some works have explored the impact of faults and non-idealities on pruned models, and have shown that pruned DNNs have a larger accuracy degradation in the presence of faults and non-idealities compared to their unpruned counterparts \cite{faults_non_ideal}\cite{faults_pruning}. However, the understanding of the impact of faults on \textit{activation-sparse} DNNs is lacking. Moreover, techniques that can alleviate the adverse effect of faults on accuracy of activation-sparse QDNNs are needed to increase their computational robustness in non-ideal settings.

In this paper, we address these critical needs by (a) extensively analyzing the performance of activation-sparse QDNNs in the presence of stuck-at and bit-flip faults and (b) proposing a training technique to enhance fault tolerance.  The key contributions of our work are as follows:
\begin{itemize}
    \item We show that an increase in activation sparsity comes at the expense of reduced fault tolerance and lower inference accuracy in QDNNs. To the best of our knowledge, this is the first work exploring the performance of \textit{activation-sparse QDNNs} in the presence of faults.  
    \item  Using the weight loss landscape visualization method in \cite{visualize}, we establish that the higher sensitivity of activation-sparse QDNNs to faults is attributed to sharper minima in their loss landscape (compared to the standard QDNNs).
    \item Based on the above finding, we propose the use of sharpness-aware quantization (SAQ) \cite{saq} (which is a variant of sharpness-aware minimization (SAM) \cite{sam} designed for QDNNs) to mitigate the impact of faults on system accuracy. 
    \item We show that the proposed method increases the inference accuracy of activation-sparse and standard QDNNs by up to 19.50\% and 15.82\% by reducing the sharpness of the weight loss landscape. This is achieved without compromising on the baseline software accuracy. 
    \item We also show that SAQ-trained activation-sparse QDNNs have higher inference accuracy than their standard counterparts trained without SAQ. This enables the design of QDNNs which are both activation-sparse and fault tolerant, optimal for edge applications. 
\end{itemize}

Figure ~\ref{fig:overview} provides an overview of our work and contributions.

\section{Background and Related Work}
\label{sec:background}

\subsection{Activation sparsity in DNNs}
Activation sparsity refers to the prevalence of a large number of zero values in DNN activations. The most common activation function in DNNs is the rectified linear unit (ReLU), which outputs a zero for every negative input, leading to a high activation sparsity. As a result, sparse storage schemes can be used to store the activations, reducing memory requirements \cite{sparse_code}. Also, the computations involving the zero-valued activations can be skipped, reducing the energy and latency of the DNNs \cite{act_sparse3}. Note that activation sparsity is dynamic in nature, which means that the number and location of the zero values vary from input to input. This property needs to taken into account while exploiting activation sparsity to reduce the computation and memory requirements. Hence, when utilized properly, a large activation sparsity can be highly beneficial for designing resource-constrained DNN accelerators. 

Due to this reason, algorithmic techniques to \textit{enhance} the activation sparsity have gained interest in recent times. These techniques are primarily based on explicitly adding a regularization term to the loss function which penalizes dense activations. For example, the work in \cite{act_sparse2} adds the $L_{1}$ norm of the activations (\(||x_{l,n}||_{1}\)) to the original loss function to incentivize sparsity:
\begin{equation}
    L_{reg}(x,w) = L_{0}(x,w) + \sum^{N}_{n=1}\sum^{L}_{l=1}\alpha_{l}||x_{l,n}||_{1}
\label{eq1}
\end{equation}

Here,$L_{reg}(x,w)$ is the new loss function to be minimized, $L_{0}(x,w)$ is the original loss function value, $L$ is the number of layers in the DNN, $N$ is the batch size and $\alpha_{l}$ is the regularization constant per layer.  By using the sparsifying properties of $L1$ regularization \cite{l1book}, the work in \cite{act_sparse2} demonstrates up to 60\% increase in sparsity with negligible loss in inference accuracy for image classification. The work in \cite{act_sparse1} uses a Hoyer sparsity metric based regularizer and a variant of the ReLU activation function to boost the activation sparsity of DNNs. It should be noted that these techniques are not equivalent to "pruning" the activations, analogous to weight pruning \cite{pruning1} \cite{pruning2}. While pruning permanently sets the parameters to zero for all inputs, the activation sparsification techniques do not remove any of the activations permanently. Rather, they ensure that a low percentage of activations are non-zero for all inputs \textit{on an average}. 

In this paper, we use the $L_{1}$ regularization based approach to train activation-sparse QDNNs due to its intuitive appeal and ease of implementation. We will refer to the QDNNs trained without $L1$ activation regularization as standard QDNNs and the ones trained with $L_{1}$ activation regularization as activation-sparse (AS) QDNNs. 

\subsection{Memory faults in DNN accelerators}
Aggressive scaling and the exploration of new memory technologies have made the study of memory faults more important than ever. All the popular memories (SRAMs, ReRAMs, FeFETs, etc.) used for DNN accelerators are impacted by faults. Faults corrupt a percentage of the stored values, leading to inaccurate computation of multiply-and-accumulate (the most common kernel in DNNs). This, in turn, degrades the inference accuracy. Two common memory faults which plague DNN accelerators are bit-flip faults and stuck-at faults. Stuck-at one (SA1) and stuck-at zero (SA0) faults occur when the value of the bitcell \textit{unalterably} gets fixed at '1' and '0' respectively  SA1 and SA0 faults are usually caused by fabrication defects \cite{stuck_at_faults} and limited endurance. Bit-flip faults occur when the bitcell value gets flipped (from '0' to '1' or vice versa). They can be permanent or transient and are caused by phenomena such as half-select read disturbance and alpha particle strikes.  

Understanding the impact of memory faults on DNN performance has attracted attention in recent times. Works such as \cite{graceless} have shown that a single targeted bit flip can significantly degrade the performance of floating point DNNs. Even QDNNs have been shown to be quite vulnerable to stuck-at faults and bit-flip faults \cite{dnn_faults1}\cite{dnn_faults2}. Some works have also analyzed the performance of weight sparse (pruned) networks in the presence of faults \cite{faults_pruning}, circuit non-idealities and variations \cite{faults_non_ideal},showing a larger vulnerability of pruned DNNs to faults and other non-idealities than their unpruned counterparts. However, to the best of our knowledge, no work has analyzed the impact of faults on activation-sparse QDNNs. 

With regard to the fault mitigation, a plethora of hardware and algorithmic solutions have been analyzed. Hardware-based techniques are generally based on adding redundancy by identifying and duplicating the critical portions of the DNN. \cite{hardware_mitigation1} utilizes exhaustive fault injection to identify the vulnerable parts of the DNNs and replicates those portions to reduce the impact of faults. \cite{hardware_mitigation2} adds a spare neuron to the DNN, which can be configured to act as any of the other neurons in case they turn faulty. From the algorithmic perspective, fault mitigation strategies include using error-correcting codes, fault-aware retraining and fault-tolerant training. Works like \cite{software_mitigation1} use error-correcting output codes to reduce the sensitivity to variations and SAFs. Fault injection during training \cite{software_mitigation2} is also promising but may not work well when the system is prone to more than one kind of faults. Fault-aware retraining is based on the online detection of faults and then retraining the non-faulty parameters to recover the accuracy of the DNN \cite{fault_aware_retraining}. This method furnishes promising results but may be challenging to implement for edge applications in which  online fault detection may not be supported.  

In this work, we will extensively investigate the impact of stuck-at and bit-flip faults on the performance of standard and AS QDNNs.We will also propose a fault mitigation strategy which utilizes SAQ to flatten the weight loss landscape of the QDNN and make its parameters less sensitive to perturbations and hence, more fault-tolerant. Our proposed technique can be used in conjunction with hardware based fault tolerance techniques and provides protection from multiple kinds of faults, as will be discussed subsequently.

%
%
%

\subsection{Sharpness-Aware Quantization (SAQ)}
The sharpness of the minima of the loss landscape that a DNN converges to during training has a key impact on its generalization capability \cite{generalization1}\cite{generalization2}. Both theoretically and empirically, it has been shown that convergence to a flat minima improves generalization. Hence, sharpness-aware training schemes, such as sharpness-aware minimization (SAM) have been explored in \cite{sam}. SAM simultaneously minimizes the loss value and the sharpness of the weight loss landscape to learn optima which have uniformly low loss values in their neighbourhood. DNNs trained with SAM have been shown to achieve state-of-the-art accuracies for several benchmark datasets and models. However, SAM is not as effective in QDNNs as it does not account for the quantization of weights. Sharpness-aware quantization (SAQ) is a variant of SAM designed for QDNNs \cite{saq}. SAQ simultaneously minimizes the loss value and flattens the loss curvature near the minima by minimizing the loss function with adversarially perturbed quantized weights. This leads to better generalization and improved accuracy of QDNNs.  
In this work, we advance the application of SAQ to train fault-tolerant QDNNs, based on the intuition that QDNNs converged at flatter minima will be less sensitive to the weight perturbations caused by faults.

\section{Impact of Faults on Activation-Sparse QDNNs}
\label{sec:impact}
\subsection{Experimental Framework}
To investigate the impact of faults on both regular and activation-sparse (AS) QDNNs, we utilize the following methodology. First, two 4-bit QDNNs, i.e. LeNet 5 \cite{lenet} with the FashionMNIST (FMNIST) dataset \cite{fmnist} and ResNet 18 \cite{resnet} with the CIFAR-10 dataset \cite{cifar}  are trained using the quantization framework of \cite{google_quant} in Pytorch. The AS QDNNs are trained by combining the $L1$ activation regularization described in \cite{act_sparse2} and the quantization framework. The standard and AS QDNNs are trained to have nearly equal ($\leq 0.5\%$) inference accuracies to ensure a fair comparison. 

Once the QDNNs are trained, two types of faults viz. bit-flip faults and stuck-at faults (SA0 and SA1 individually) are injected. The faults are distributed randomly and uniformly across each of the models. Monte Carlo based fault injection experiments are performed for LeNet 5 and ResNet 18 respectively. The mean value of the inference accuracy is analyzed for different fault rates and types of faults (see Figure ~\ref{fig:total_saq}). We perform experiments for fault rates 1\% - 5\% and 0.5\% - 3\% for the LeNet 5 and ResNet 18 QDNNs respectively. Fault rates up to 3\% were used for our analysis of ResNet 18 QDNNs because higher fault rates led to severe accuracy drops that would deem the QDNNs unusable.

To gain further insight into the difference in their fault tolerance, we visualize the weight loss landscape for LeNet 5 standard and AS QDNNs. Various works have utilized loss landscape visualization to enhance understanding of model performance and behaviour, showing correlation between generalization and the shape of minima for DNNs. To the best of our knowledge, this is the first work which investigates the shape of the minima with respect to the QDNN's fault tolerance. We use the normalization-based visualization technique described in \cite{visualize} to generate the weight loss landscapes for LeNet 5 standard and AS QDNNs.

\subsection{Results}
\subsubsection{Activation sparsity}
To understand the impact of faults on activation sparse QDNNs, let us first discuss the gain in activation sparsity obtained due to $L1$ activation regularization. Figures ~\ref{fig:sparsity}a and ~\ref{fig:sparsity}b show the percentage of zero activations (averaged over all layers and test inputs) for LeNet 5 and ResNet 18 QDNNs, respectively. For LeNet 5, training with $L1$ regularization increases the activation sparsity by 95.34\% (from 43\% for standard training to 84\% for $L1$ regularization ). For ResNet-18, the activation sparsity increases by 41.51\% (from 53\% to 75\%). The difference in the activation sparsity increase between LeNet 5 and ResNet-18 is attributed to the higher workload complexity for ResNet 18 (image classification on the CIFAR-10 dataset)  compared to LeNet-5 (FMNIST dataset). As a result, ResNet-18 requires more nonzero activations on an average to sustain high accuracy. 

\begin{figure}[t!]
\centering
  \includegraphics[width = \linewidth]{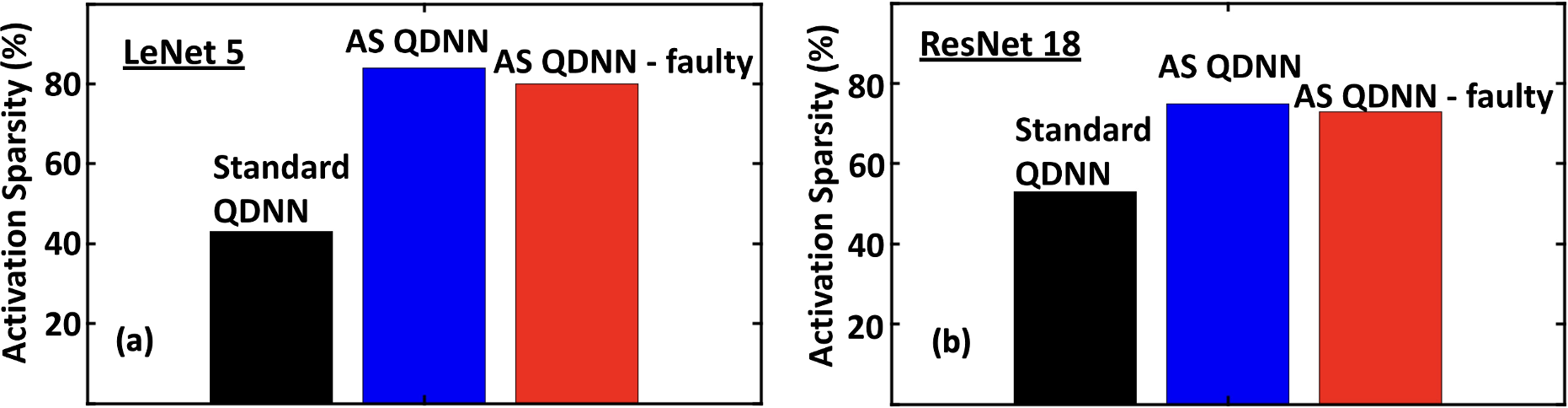}
  \caption{The activation sparsities of (a) LeNet 5 and (b) ResNet 18 standard QDNNs and activation sparse (AS) QDNNs in both fault-free and faulty settings. The activation sparsity of LeNet 5 and ResNet 18 AS QDNNs is 95.34\% and 41.51\% higher than their standard counterparts, and is sustained in faulty environments.}
  \label{fig:sparsity}
\end{figure}

We further analyze how the sparsity of activation-sparse QDNNs are impacted in faulty settings. Based on our fault injection analysis, we observe a negligible change in activation sparsity. In the worst case (5\% and 3\% bit flip fault rate respectively), LeNet 5 and ResNet 18 show a ~4\% and ~2\% reduction in sparsity, respectively. This signifies that the gain in activation sparsity is sustained even in the presence of faults.

\subsubsection{Impact of faults on Inference Accuracy}
The dashed lines in  Figures ~\ref{fig:saq_lenet_bit_flip}-~\ref{fig:saq_resnet_sa1} show the impact of bit-flip, SA0 and SA1 faults on standard and AS LeNet 5 and ResNet 18 QDNNs. We observe that the AS QDNNs have lower inference accuracy in the presence of faults compared to the standard QDNNs. For the bit-flip faults, the accuracy of LeNet 5 and ResNet 18 AS QDNNs is 2.40\% to 11.13\% and 0.52\% to 8.00\% lower (absolute difference in accuracy) than their equivalent standard QDNNs respectively.  This trend holds for SA0 and SA1 faults, albeit with lower accuracy degradation than bit-flip faults. This is due to the fact that stuck-at faults can potentially get masked and hence, are less severe than bit-flip faults. These results signify that both LeNet 5 and ResNet 18 AS QDNNs have higher accuracy degradation in the presence of faults than their standard QDNN counterparts, implying that increased activation sparsity comes at the price of reduced tolerance to faults. 

It should be noted that the accuracy degradation for ResNet 18 AS QDNNs (up to 8.00\%) is more than that of LeNet 5 AS QDNNs (up to 5.56\%) for the same fault rate (up to 3\%). This can be attributed to the higher workload complexity for ResNet 18 on CIFAR-10 compared to LeNet-5 on FMNIST, making it more susceptible to accuracy degradation due to faults.     

\begin{figure}[t!]
\centering
  \includegraphics[width = \linewidth]{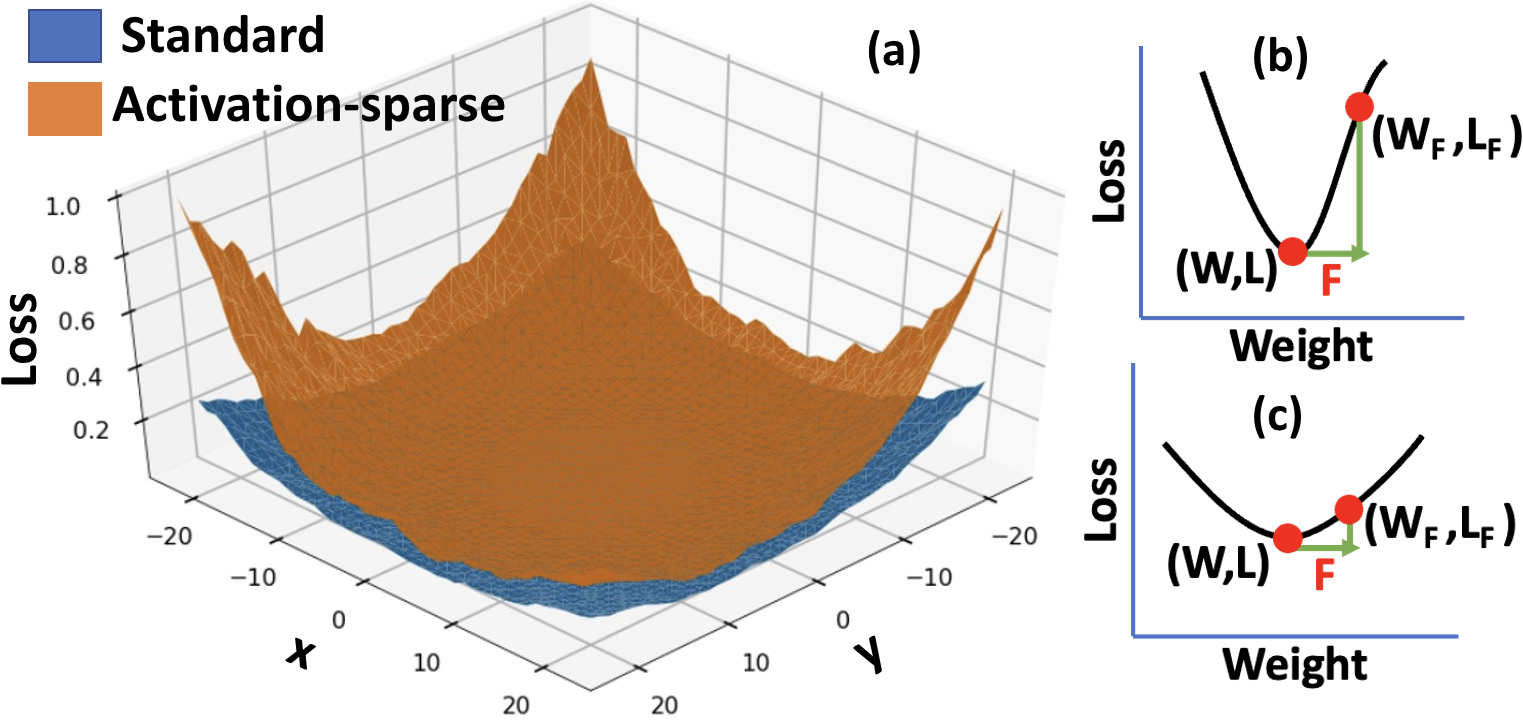}
  \caption{(a) The weight loss landscape of the standard and activation sparse (AS) LeNet 5 QDNNs visualized using the technique in \cite{visualize}. x and y are normalized random directions. (b) and (c) illustrate the impact of a fault (F) on the loss value in loss landscape with (b) sharp and (c) flat minima. The fault causes a larger change in the loss value in the sharp minima case.}
  \label{fig:sharp}
\end{figure}

\begin{figure*}[t!]
\centering
\begin{subfigure}{0.32\textwidth}
  \includegraphics[width=\linewidth]{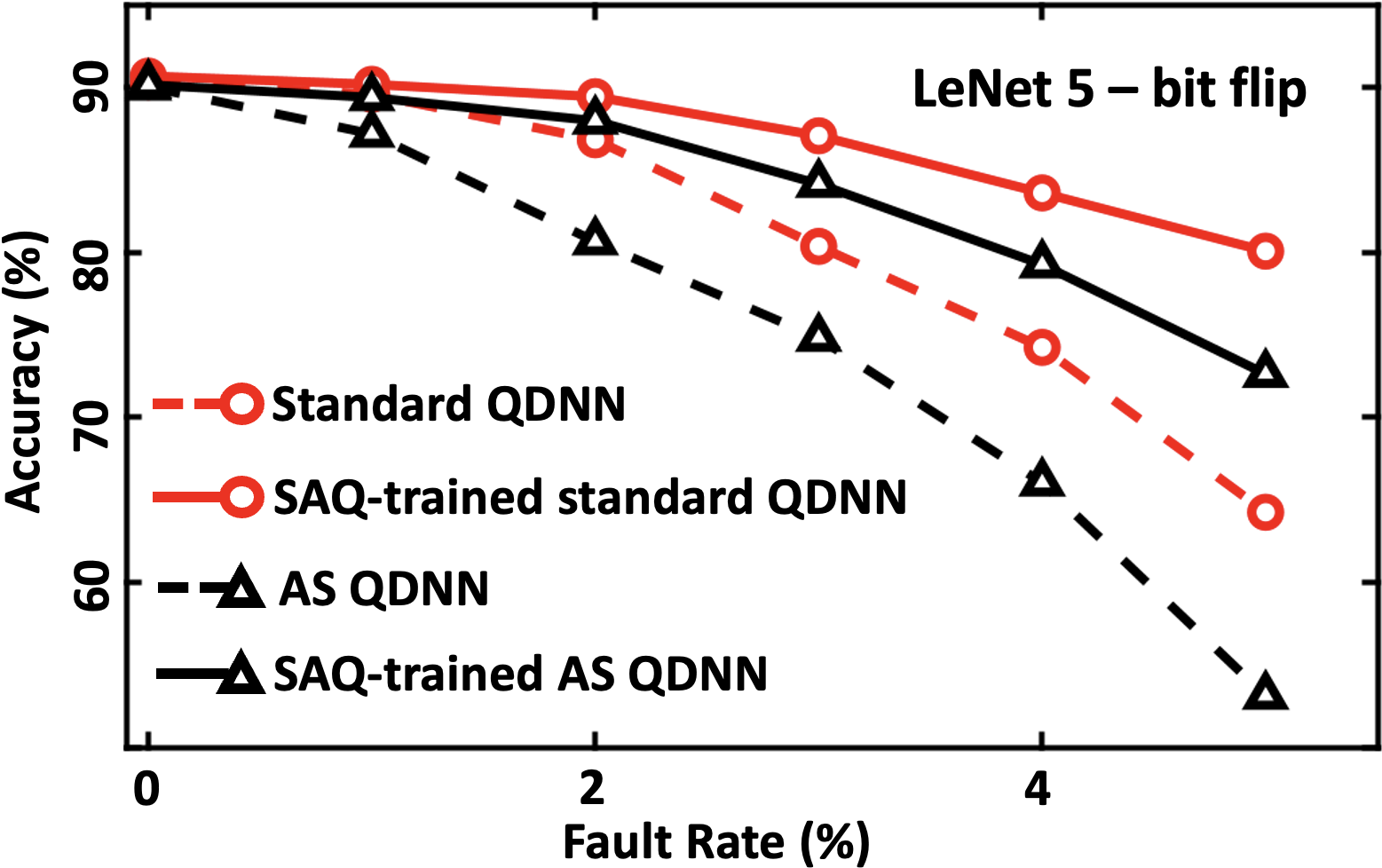}
  \caption{Bit-flip faults on LeNet 5 QDNNs}
  \label{fig:saq_lenet_bit_flip}
\end{subfigure}%
~
\begin{subfigure}{0.32\textwidth}
  \includegraphics[width=\linewidth]{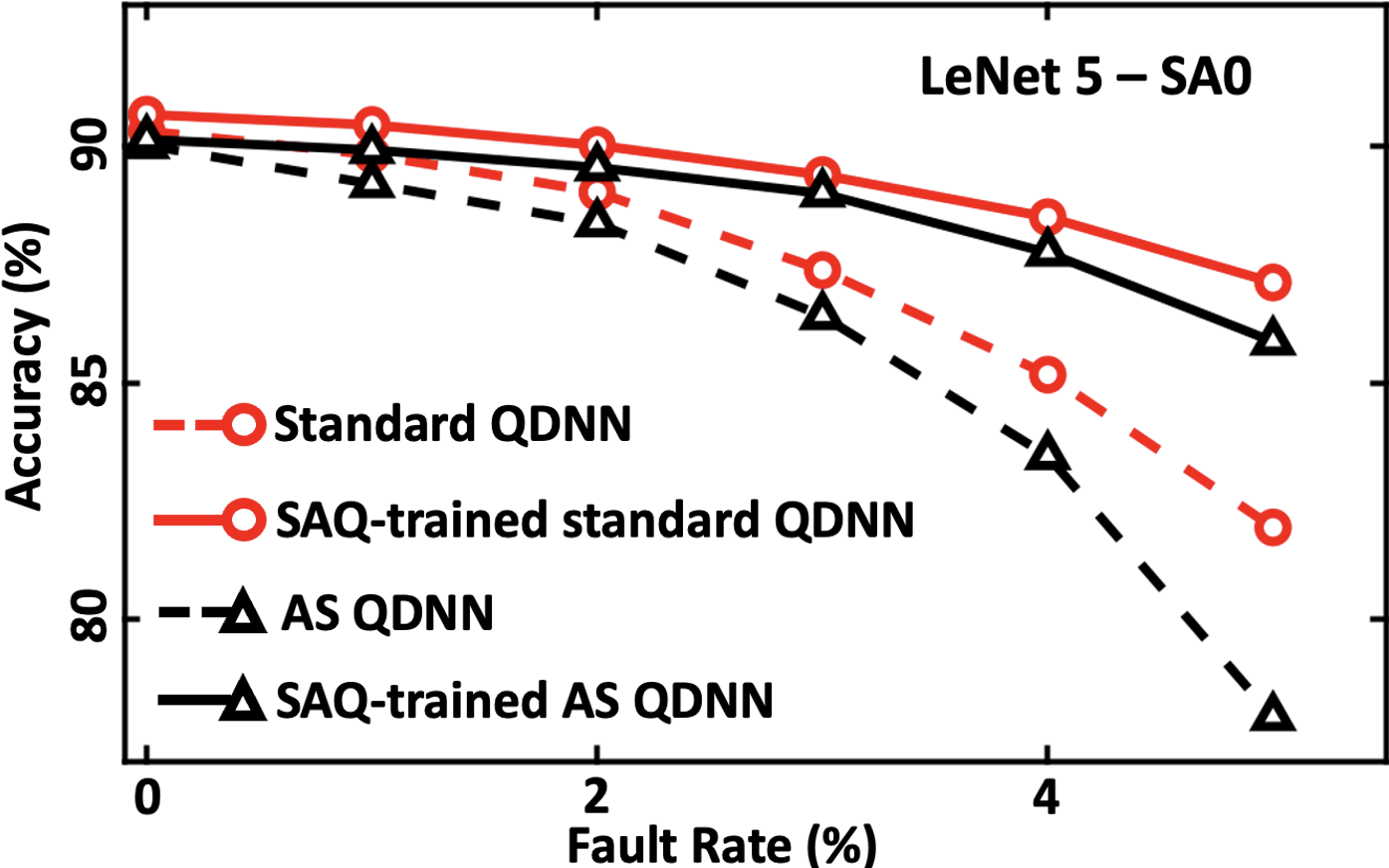}
  \caption{SA0 faults on LeNet 5 QDNNs}
  \label{fig:saq_lenet_sa0}
\end{subfigure}%
~
\begin{subfigure}{0.32\textwidth}
  \includegraphics[width=\linewidth]{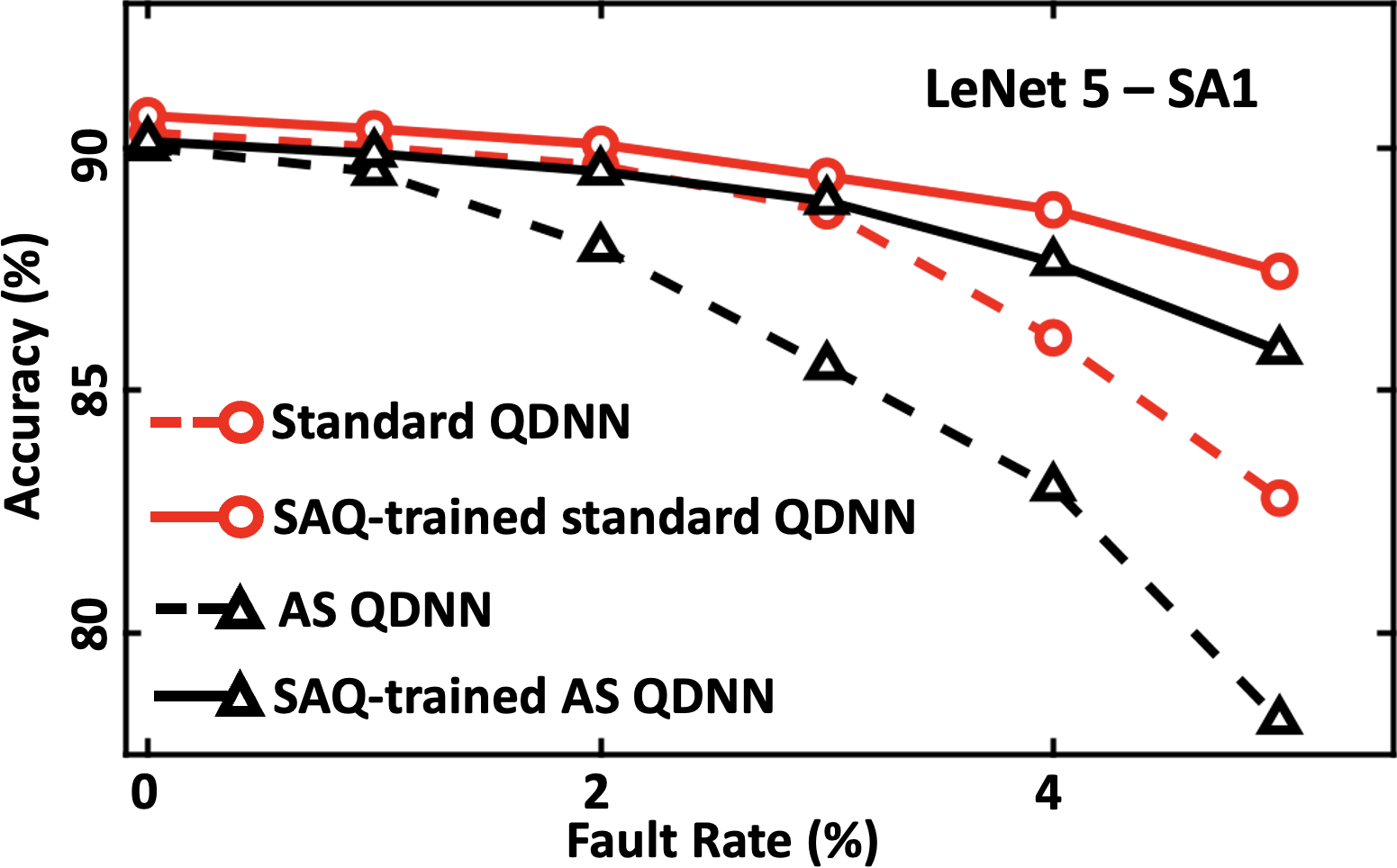}
  \caption{SA1 faults on LeNet 5 QDNNs}
  \label{fig:saq_lenet_sa1}
\end{subfigure}%
~
\vspace{4mm}

\begin{subfigure}{0.32\textwidth}
  \includegraphics[width=\linewidth]{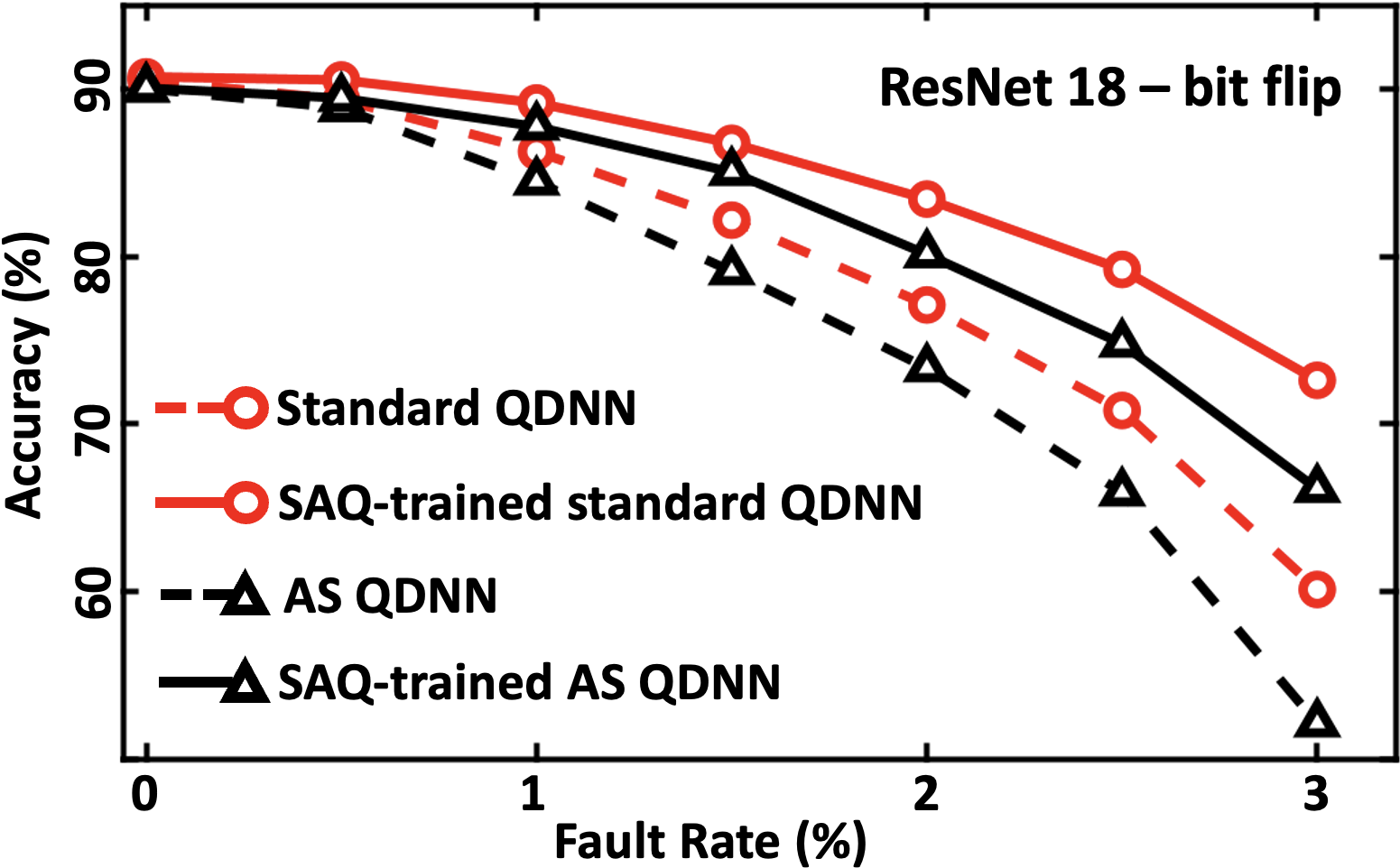}
  \caption{Bit-flip faults on ResNet 18 QDNNs}
  \label{fig:saq_resnet_bit_flip}
\end{subfigure}%
~
\begin{subfigure}{0.32\textwidth}
  \includegraphics[width=\linewidth]{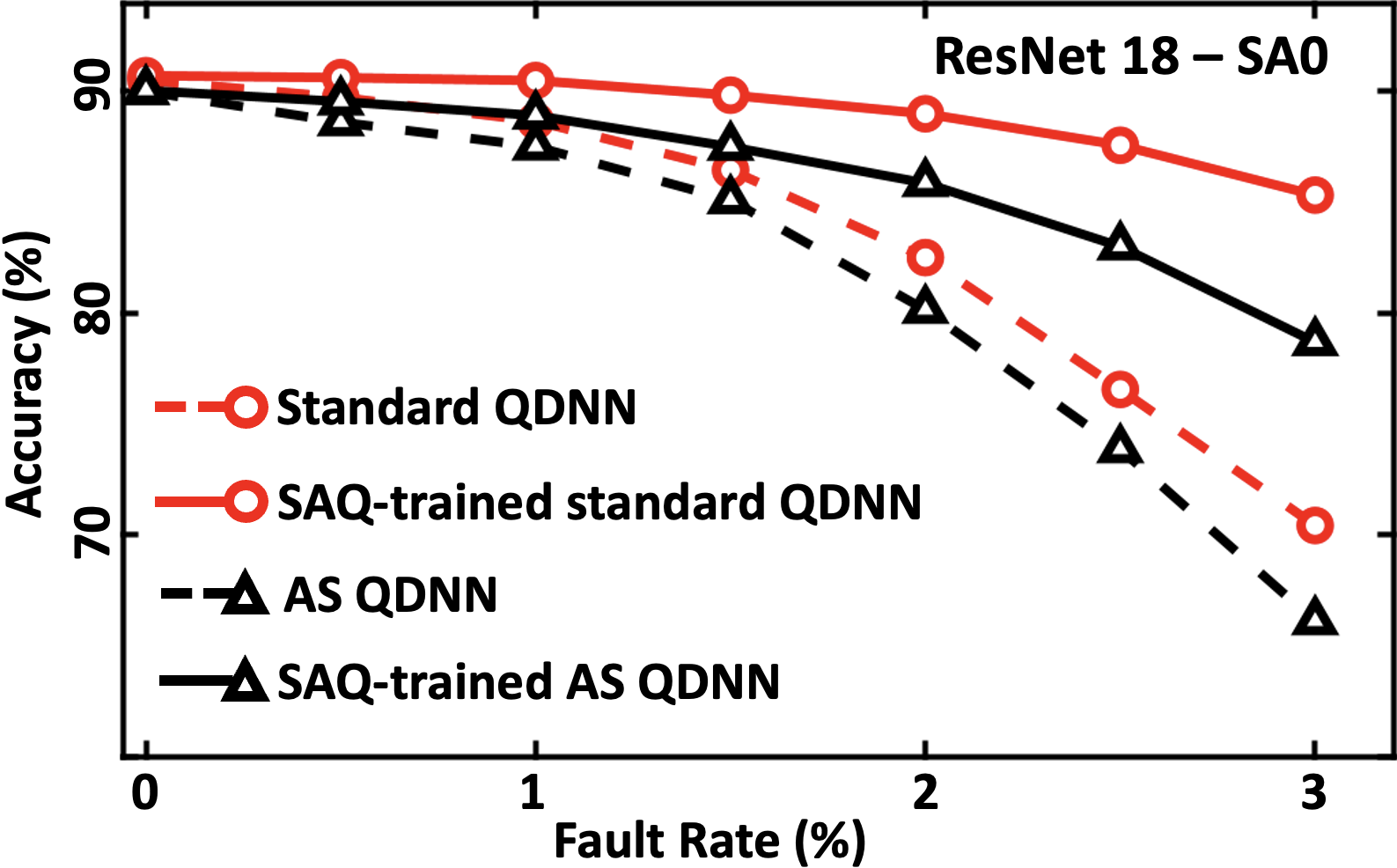}
  \caption{SA0 faults on ResNet 18 QDNNs}
  \label{fig:saq_resnet_sa0}
\end{subfigure}%
~
\begin{subfigure}{0.32\textwidth}
  \includegraphics[width=\linewidth]{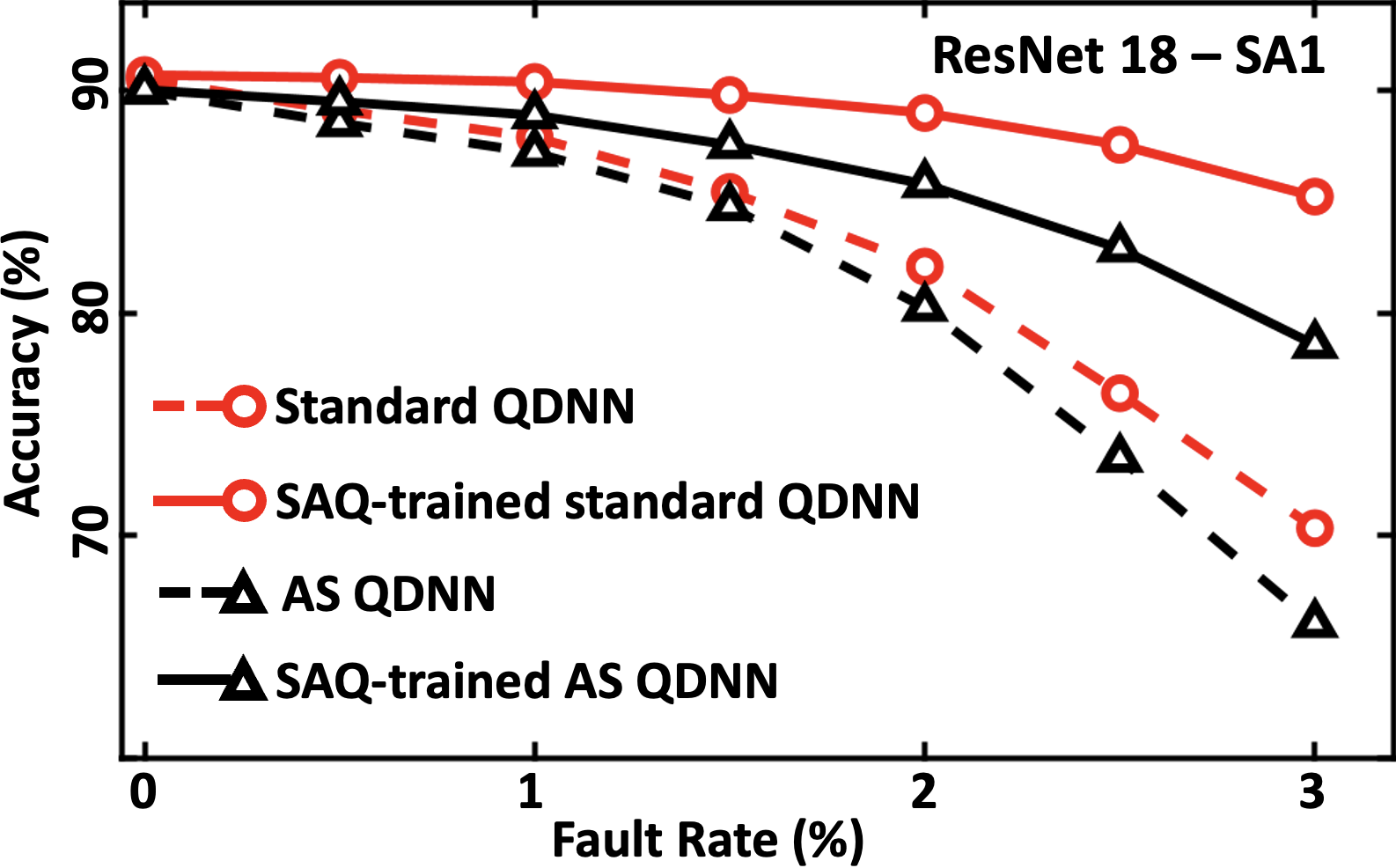}
  \caption{SA1 faults on ResNet 18 QDNNs}
  \label{fig:saq_resnet_sa1}
\end{subfigure}%
~
\vspace{4mm}
\caption{Comparison of the impact on classification accuracy for different fault scenarios for both SAQ trained and conventionally trained standard and activation-sparse (AS) QDNNs. It can be seen that AS QDNNs have lesser fault tolerance than their standard counterparts. Also, SAQ-trained QDNNs display higher fault tolerance than their convnentionally trained equivalents. }
\label{fig:total_saq}

\end{figure*}
\subsubsection{Weight Loss Landscape visualization}
Figure ~\ref{fig:sharp}a shows the weight loss landscape of the standard and AS LeNet 5 QDNN (created using \cite{visualize}). Intuitively, it is understood that a QDNN with a sharper minima would incur a larger change in its loss function value for some perturbation to its weights than a QDNN with a flat minima, as shown in Figure \ref{fig:sharp}b and c. It can be seen that the AS QDNN converges at a sharper minima, which explains its degraded fault tolerance empirically observed in Section \ref{sec:impact}. Thus, the increased activation sparsity due to the $L1$ activation regularization is correlated with the increase in the sharpness of the minima and reduced fault tolerance.

To sum up, we conclude that additional activation sparsity, which reduces the latency and energy consumption of optimized DNN accelerators, comes at the cost of degraded performance in the presence of faults. Motivated by the need to enhance the fault tolerance of AS QDNNs, we now propose how the impact of faults can be alleviated by flattening the weight loss landscape via sharpness-aware quantization (SAQ) based training. 

\section{Mitigation Strategy: SAQ}
\label{sec:mitigation}
In this section, we propose and study the performance of our SAQ training-based fault mitigation strategy for reducing the accuracy degradation in both standard QDNNs and activation-sparse (AS) QDNNs in the presence of bit-flip and stuck-at faults. We perform our experiments using the framework described in Section \ref{sec:impact}. 

\subsection{SAQ-based fault mitigation strategy}
Our fault mitigation technique is based on flattening the weight loss landscape of the QDNNs to make the weights less sensitive to faults. To implement and evaluate our proposed fault mitigation strategy, we utilize SAQ to train both standard and AS QDNNs. 

SAQ concurrently minimizes the loss value and the loss sharpness by solving the following min-max optimization problem:
\begin{multline}
   \underset{w}{\min} \underset{||\epsilon||_{2} \leq \rho}{\max}((L_{S}(Q_{w}(w,b)) + \epsilon) - L_{S}(Q_{w}(w,b))) \\
   + L_{S}(Q_{w}(w,b)) + \frac{\lambda}{2}||w||^{2}_{2} 
\label{eq2}
\end{multline}

The first term in the equation defines the sharpness metric, which is the maximum change in the loss value for some weight perturbation $\epsilon$. The second term is the loss function itself and the third term is the standard $L_{2}$ regularization term. Also, the perturbation $\epsilon$ is an adversarial one and is chosen such that the maximum sharpness is minimized. $\epsilon$ is given be the following equation and can have a maximum value of $\rho$. It has the value:
\begin{equation}
\epsilon \simeq \hat{\epsilon} = \rho\frac{\nabla_{Q_{w}(w,b)}L_{S}(Q_{w}(w,b))}{||\nabla_{Q_{w}(w,b)}L_{S}(Q_{w}(w,b))||_{2}}   
\label{eq3}
\end{equation}

where $\nabla_{Q_{w}(w,b)}L_{S}(Q_{w}(w,b))$ is the gradient of the loss function with respect to the quantized weights. $\epsilon$ is estimated using a forward and backward pass through the QDNN.

It is important to note that one epoch of SAQ training takes the same time as two epochs of conventional training. Hence for a fair comparison, we train the SAQ-based QDNNs with half the number of epochs compared to the conventionally trained QDNNs. The $\rho$ hyperparameter for the SAQ-based QDNNs is chosen such that the fault-free accuracy (fault rate = 0\%) of the QDNNs is maximized. Using this framework, we analyse the effectiveness of SAQ in increasing the fault tolerance for both standard QDNNs and AS QDNNs.

\subsection{Results}
\subsubsection{Standard QDNNs}
Figure \ref{fig:total_saq} shows the performance of LeNet 5 and ResNet 18 standard QDNNs trained with SAQ compared to the performance of those conventionally trained in the presence of faults. It can be seen that SAQ-trained standard QDNNs have superior fault tolerance to their conventionally trained equivalents. For LeNet 5 with fault rates of 0\% - 5\%, the SAQ trained model has a 0.35\% - 15.82\% , 0.35\% - 5.18\% and 0.35\% - 4.68\% higher inference accuracy for bit-flip, SA0 and SA1 faults, respectively. Even at a fault rate of 0\% (fault-free environment), SAQ-trained standard QDNNs outperform conventionally trained standard QDNNs, which is consistent with the results in \cite{sam} and \cite{saq}. Similarly, for ResNet 18, the SAQ-trained standard QDNNs  display better inference accuracies compared to conventionally trained standard QDNNs in both fault-free and faulty settings. The accuracy improvements  for bit-flip, SA0 and SA1 faults are  0.21\% - 12.48\%, 0.21\% - 14.89\% and 0.21\% - 14.92\%, respectively.    

\subsubsection{Activation-sparse (AS) QDNNs}
The comparison of the inference accuracies of SAQ-trained and conventionally trained LeNet 5 and ResNet 18 AS QDNNs can be seen in Figure \ref{fig:total_saq}. The $L1$ regularization constant for both the SAQ-trained and conventionally trained QDNNs is kept the same, so that the activation sparsity in both of them is nearly equal (value in figures \ref{fig:sparsity}a and \ref{fig:sparsity}b).

In trend with standard QDNNs, it is observed that SAQ-trained AS QDNNs have higher inference accuracy than their conventionally trained counterparts. For LeNet 5, the SAQ trained AS QDNN has a 0.12\% - 19.50\% , 0.12\% - 7.94\% and 0.12\% - 7.63\% higher inference accuracy for bit-flip, SA0 and SA1 faults respectively, For ResNet 18, the SAQ-trained AS QDNN again displays superior inference accuracies in both fault-free and faulty settings, with  0.04\% - 14.00\%, 0.04\% - 12.59\% and 0.04\% - 12.56\% higher accuracy than the conventionally trained model for bit-flip, SA0 and SA1 faults respectively. 

An interesting point to note here is that activation-sparse (AS) QDNNs trained with SAQ have better performance in the presence of faults than standard QDNNs without SAQ. For example, our experiments show that the SAQ-trained LeNet 5 AS QDNN has a 8.37\% higher accuracy than the conventionally trained LeNet 5 standard QDNN at fault rate = 5\%. Thus, SAQ-trained AS QDNNs have both the benefits of superior fault tolerance \textit{and} increased activation sparsity (which leads to lower latency and energy consumption), which makes them highly suitable for edge applications. 

Lastly, we see that training with SAQ limits the accuracy degradation caused by enhancing the activation sparsity. For both LeNet 5 (fault rate = 5\%) and ResNet 18 (fault rate = 3\%), the respective AS QDNNs trained with SAQ have 7.45\% and 6.49\% lower inference accuracy than the SAQ-trained standard QDNNs, whereas the conventionally trained AS QDNNs have 11.13\% and 8.00\% lower accuracy. Thus, training with SAQ mitigates the adverse impact that activation sparsity augmentation has on the fault tolerance.

\section{Conclusion}
\label{sec:conclusion}

We explored the performance of activation-sparse (AS) QDNNs in the presence of faults and proposed a mitigation strategy to reduce the impact of faults on the accuracy of the QDNNs. Through our experiments, in which we uniformly inject different kinds of faults in LeNet 5 and ResNet 18 QDNNs, we show that the increase in activation sparsity comes at the price of degraded inference accuracy in the presence of faults, with activation-sparse QDNNs showing up to 11.13\% lower accuracy than standard QDNNs. To understand the reason for this reduced fault tolerance, we visualize the weight loss landscape for the standard and AS QDNNs and show that the AS QDNN has a sharper minima leading to lower fault tolerance. Based on this observation, we propose the flattening of the loss landscape for enhanced fault tolerance by utilizing sharpness-aware quantization (SAQ) based training. Our results show that AS and standard QDNNs trained with SAQ have upto 19.50\% and 15.82\% higher accuracy than their conventionally trained counterparts. We also observe that SAQ-trained AS QDNNs have higher inference accuracy than conventionally trained standard QDNNs, enabling QDNNs which are not only activation-sparse but also fault tolerant. Thus, SAQ-trained QDNNs have enhanced fault tolerance, making them suitable for deployment in fault-prone edge scenarios.  

\bibliography{references.bib}
\bibliographystyle{IEEEtran}

\end{document}